\documentclass[final,5p,times,twocolumn]{elsarticle}

\usepackage{lineno}
\usepackage{graphicx}
\usepackage{amssymb}
\usepackage{pdflscape}
\usepackage[english]{babel}
\usepackage[dvipsnames,svgnames,x11names]{xcolor}
\usepackage{booktabs,tabularx}
\usepackage{multirow}
\usepackage{subfig} 
\usepackage{hyperref}
\usepackage{amsmath}
\usepackage{commath}
\usepackage[english]{babel}
\usepackage[ruled]{algorithm2e}
\SetEndCharOfAlgoLine{}
\usepackage[leftmargin=6em,rightmargin=12em,indentfirst=false]{quoting}
\usepackage{siunitx}

\usepackage[final]{changes}
\usepackage{float}

\usepackage{lineno}
\usepackage{tikz}
\usepackage[export]{adjustbox}

\usepackage{graphicx}
\usepackage[utf8]{inputenc}
\usepackage[export]{adjustbox}
\usepackage{wrapfig}
\usepackage{dcolumn}

\makeatletter
\newcolumntype{T}[3]{>{\textfont0=\the@{#1}{#2}{#3}}c<{\DC@end}}
\makeatother

\usepackage{pgfplots}
\pgfplotsset{width=10cm,compat=1.9}

\usepackage{array}

\newcolumntype{L}[1]{>{\raggedright\let\newline\\\arraybackslash\hspace{0pt}}m{#1}}
\newcolumntype{C}[1]{>{\centering\let\newline\\\arraybackslash\hspace{0pt}}m{#1}}
\newcolumntype{R}[1]{>{\raggedleft\let\newline\\\arraybackslash\hspace{0pt}}m{#1}}

\usepackage{subfiles}

\usepackage{todonotes}

\journal{Science and Technology for the Built Environment}
\begin{document}
	
\begin{frontmatter}

\title{Limitations of machine learning for building energy prediction: ASHRAE Great Energy Predictor III Kaggle competition error analysis}

\author{Clayton Miller\,$^{1,*}$, Bianca Picchetti\,$^{1}$, Chun Fu\,$^{1}$, Jovan Pantelic\,$^{2,3}$}

\address{$^{1}$Department of the Built Environment, College of Design and Engineering, National University of Singapore (NUS), Singapore}
\address{$^{2}$Department of Biosystems, KU Leuven, Belgium}
\address{$^{3}$Well Living Lab, Delos Living LLC, USA}
\address{$^*$Corresponding Author: clayton@nus.edus.sg, +65 81602452}

\begin{abstract}
Research is needed to explore the limitations and potential for improvement of machine learning for building energy prediction. With this aim, the ASHRAE Great Energy Predictor III (GEPIII) Kaggle competition was launched in 2019. This effort was the largest building energy meter machine learning competition of its kind, with 4,370 participants who submitted 39,403 predictions. The test data set included two years of hourly whole building readings from 2,380 meters in 1,448 buildings at 16 locations. This paper analyzes the various sources and types of residual model error from an aggregation of the competition's top 50 solutions. This analysis reveals the limitations for machine learning using the standard model inputs of historical meter, weather, and basic building metadata. The errors are classified according to timeframe, behavior, magnitude, and incidence in single buildings or across a campus. The results show machine learning models have errors within a range of acceptability (RMSLE$_{scaled}$ $=<$ 0.1) on 79.1\% of the test data. Lower magnitude (in-range) model errors (0.1 $<$ RMSLE$_{scaled}$ $=<$ 0.3) occur in 16.1\% of the test data. These errors could be remedied using innovative training data from onsite and web-based sources. Higher magnitude (out-of-range) errors (RMSLE$_{scaled}$ $>$ 0.3) occur in 4.8\% of the test data and are unlikely to be accurately predicted. 
\end{abstract}


\end{frontmatter}


\section{Introduction}
Building energy prediction using machine learning models has been around for decades and has become important for the evaluation of performance in the context of retrofits~\citep{Deb2021-py, Grillone2020-fm}, measurement and verification~\citep{Gallagher2018-ji, Amasyali2018-wj}, renewable energy integration~\citep{Perera2014-nq}, systems control~\citep{Brandi2020-ea, Fan2017-ac}, fault detection~\citep{Gunay2017-ke}, residential energy use~\citep{Jahani2020-qb}, and urban-scale energy modeling~\citep{Roth2020-bb, Nutkiewicz2018-dk}. A recent text mining-driven review of 30,000 building energy-related data science publications shows that there has been rapid growth in the last ten years in techniques and applications~\citep{Abdelrahman2021-bk}. This rapid expansion of the field is creating a myriad of prominent techniques using deep learning~\citep{Nichiforov2018-ne, Fan2017-ac, Wang2020-ky}, embedded online systems~\citep{Nichiforov2019-jh}, sequence learning~\citep{Nichiforov2019-ky}, transfer learning~\citep{Li2021-wb, Fan2020-ok}, neural networks~\citep{Nichiforov2017-vh, Li2021-wb}, bayesian probabilistic forecasting~\citep{Roth2021-be}, and gradient boosting trees~\citep{Touzani2018-sa, Ahmad2017-jz}. 

Despite this expansion and innovation, there has been a lack of firm understanding of which techniques, tools, and models are the most accurate, fastest, or easiest-to-use specifically for building energy consumption prediction. Benchmarking in the energy prediction context would help solve this problem by applying various methods on the same large data set, making the comparison of results meaningful~\citep{Miller2019-sg}. Several initial studies have created environments to apply energy prediction techniques across a large group of buildings~\citep{Granderson2016-wq, Granderson2015-ms, Granderson2014-xl, Granderson2017-lm}. To further expand both the data set and talent pool of machine learning experts, the Great Energy Predictor III (GEPIII) competition was planned and executed in 2019 as a means of crowdsourcing thousands of prediction solutions and determining the most effective in a competitive environment~\citep{Miller2020-qw}. GEPIII was the largest exercise in benchmarking machine learning methods for this context in terms of size of data set, number of compared techniques, and amount of open-source content shared with the community.

\subsection{What are the limitations of machine learning for energy prediction when the best models, techniques, and experts are involved?}
GEPIII brought together the combined efforts of 4,370 participants, and the top solutions can be considered near-optimal for the long-term prediction context used. These solutions have empirically found the best combination of preprocessing, feature selection, model types, and post-processing strategies for this scenario. Large ensembles of mainly gradient boosting trees with significant preprocessing of the training data were found to be the best solutions for this application~\citep{Miller2020-qw}. The areas of improvement that remain are those caused by the limitations due to the inability to predict behavior that was not observed previously and, therefore, not used to train the machine learning models. The key \emph{limitation} in this sense is that machine learning models cannot predict behavior that is not contained in the training data. Therefore, the \emph{potential} is to quantify and classify these situations to help pinpoint new sources of data that can supplement the process and improve performance in those scenarios. 

\subsection{Previous and related work}
Few studies exist that explore the sources of error for building performance prediction. In the physics-based modeling realm, the quantification of error has been heavily focused on the capabilities of the simulation model specification process to capture reality accurately. This field explores \emph{uncertainty analysis} identify the risk that assumptions made in this process are not precise or accurate enough~\citep{Tian2018-ur}. In data-driven modeling, quantification of modeling error has been explored in the context of measurement and verification~\citep{Reddy1998-ft}. Granderson et al. empirically investigated the ten modeling frameworks against 537 building meters to show that error was within a reasonable range compared to commonly-used standards~\citep{Granderson2016-wq}. This study found that the median coefficient of variation of the root mean squared error (CVRSME) was less than 25\% for various training data lengths (6 and 12 months).

Other domains have explored the use of extensive reviews or benchmarking exercises like modeling competitions to quantify error. For example, the medical field has characterized what causes errors for oncology~\citep{Jarrett2019-hs} and cardiovascular medicine~\citep{Shameer2018-bc}. There is a comprehensive analysis of categories of limitation for machine learning applied to social objectives~\citep{Malik2020-si}. Several studies specifically analyzed the results of a Kaggle competition to characterize modeling error and its lessons~\citep{Van_Aken2018-yb, Mangal2016-qi}. Other studies created their own large-scale benchmarking analysis similar to Kaggle~\citep{Seifert2020-xn}.

\subsection{Objectives and novelty}
To build upon the momentum of the GEPIII competition, this research seeks to quantify and classify the energy use behavior that the contestants could not predict even with the most advanced machine learning techniques at their disposal. The winning solutions were restrained from perfectly predicting the ground truth due to limitations in the training data and capabilities of machine learning. This paper builds upon this intent in the following ways:
\begin{itemize}
\item The method classifies sources of time-series prediction error for building energy prediction according to magnitude, shape, and number of coincidental occurrences across a collection of buildings.
\item The breadth of data and prediction submissions create a scenario in which an error analysis from this competition is the largest of its kind. The results can show the percentage of the test data that belongs to the categories of \emph{easy to predict}, \emph{prediction could be enhanced}, or \emph{prediction likely impossible} for state-of-the-art machine learning methods. These results are generalizable for the building and meter types with a critical mass in this data set (offices, classrooms, laboratories, dormitories, and municipal buildings).
\item The error analysis results are used to make practical recommendations for improving general model performance. 
\end{itemize}


Towards these objectives, the paper is structured as follows. First, Section \ref{sec:methods} outlines the process of downloading and processing the prediction data from the top 50 contestants and calculating the error for each timestamp. Each prediction data point is then classified according to defined error categories to characterize the frequency and scope of various behaviors. Next in Section \ref{sec:results}, these error categories are visualized straightforwardly from some representative sites, and then aggregations are created to show high-level trends. Next, Sections \ref{sec:discussion} and \ref{ch:conclusion} provide insight into what innovative data sources could be helpful in further reducing the overall prediction error for each category. Finally, limitations to the analysis are discussed, and reproducible code/data are linked for further work.

\section{Methodology}
\label{sec:methods}
This paper aims to characterize the error of the GEPIII competition in a way that enables classification and inference of cause. These insights help understand under and over-prediction, classify the most frequent and most important errors and identify possible remedies. Thus, the methodology first covers essential details of the competition itself, the extraction of the winning prediction data from Kaggle, and the replication of error calculation. These residuals are then classified according to the nature of the error. For example, did the error occur consistently across a period, or was it intermittent and more stochastic? These error definitions provide a foundation for investigating possible causes of error and the complementary means of addressing those errors using new data.

\subsection{Overview of the competition}
The Great Energy Predictor III competition was held on the Kaggle platform from October 5 to December 19, 2019. This competition was the next generation of the Great Energy Predictor Shootout I and II that were held in the 1990s~\citep{Haberl1998-du, Kreider1994-dn, Katipamula1996-et, Haberl1996-cs, Ohlsson1994-vl}. A detailed overview of the GEPIII competition is found in a publication focused on the contestants, the winning solutions, and the fundamental insights~\citep{Miller2020-qw}. The training data set for the competition included hourly meter data from the year 2016, while the public leaderboard test data set was the year 2017 and the private leaderboard was 2018. The 2016 and 2017 data sets were opened for use in building energy prediction studies as the Building Data Genome 2 (BDG2) project~\citep{Miller2020-hc}. In addition to the meter training data, the contestants were provided with weather data from training and test data time ranges and various metadata about each building, such as primary use type, building age, system type, and the number of floors.

\subsection{Contestant submission processing}
The first step in this analysis process was downloading all the contestants' submissions from the Kaggle platform and transforming them towards characterizing errors. Some of the buildings had to be removed from this analysis due to being publicly available during the competition. Table \ref{tab:meters_included} shows the number of buildings per site and meter type included. Table \ref{tab:overview_top50} shows examples of the data processing, input features, and modeling strategies of the public solutions among 50 submissions. Although not all solutions were publicly shared, most teams had similar strategies: removing outliers and imputing missing values in preprocessing, implementing feature engineering, and ensembling prediction results from various modeling frameworks.

\begin{table}[]
\centering
\resizebox{0.45\textwidth}{!}{
\begin{tabular}{|l|l|l|l|l|}
\hline
\textbf{Site ID} & \textbf{Electricity} & \textbf{Chilled Water} & \textbf{Steam} & \textbf{Hot Water}  \\ 
\hline
0                 & 0                     & 81                      & 105             & 105                  \\\hline
1                 & 0                     & 0                       & 0               & 51                   \\\hline
2                 & 0                     & 0                       & 0               & 0                    \\\hline
3                 & 274                   & 274                     & 274             & 274                  \\\hline
4                 & 10                    & 91                      & 91              & 91                   \\\hline
5                 & 89                    & 89                      & 89              & 89                   \\\hline
6                 & 44                    & 44                      & 44              & 44                   \\\hline
7                 & 15                    & 15                      & 15              & 15                   \\\hline
8                 & 70                    & 70                      & 70              & 70                   \\\hline
9                 & 124                   & 124                     & 124             & 124                  \\\hline
10                & 30                    & 30                      & 30              & 30                   \\\hline
11                & 5                     & 5                       & 5               & 5                    \\\hline
12                & 36                    & 36                      & 36              & 36                   \\\hline
13                & 154                   & 154                     & 154             & 154                  \\\hline
14                & 102                   & 102                     & 102             & 102                  \\\hline
15                & 0                     & 0                       & 0               & 0   
\\
\hline              
\end{tabular}}
\caption{Number of buildings included in the analysis per kind of meter and site.}
\label{tab:meters_included}
\end{table}

\begin{table*}[]
\centering
\resizebox{0.9\textwidth}{!}{
\begin{tabular}{|l|l|l|l|l|l|l|}
\hline
\textbf{Rank} & \textbf{Team} & \textbf{Score} & \textbf{Preprocess} & \textbf{Features} & \textbf{Modeling} & \textbf{Postprocess} \\ \hline
1 & \begin{tabular}[c]{@{}l@{}}Matthew Motoki and \\ Isamu Yamashita (Isamu and Matt)\end{tabular} & 1.231 & \begin{tabular}[c]{@{}l@{}}Removed anomalies \\ in meter data and \\ imputed missing \\ values in weather data\end{tabular} & \begin{tabular}[c]{@{}l@{}}28 features,   Extensively \\ focused on feature engineering \\ and selected\end{tabular} & \begin{tabular}[c]{@{}l@{}}LightGBM, CatBoost, and \\ MLP models trained on \\ different subsets of the \\ training and   public data\end{tabular} & \begin{tabular}[c]{@{}l@{}}Ensembled the model \\ predictions using weighted \\ generalized mean.\end{tabular} \\ \hline
2 & \begin{tabular}[c]{@{}l@{}}Rohan Rao, Anton   Isakin, \\ Yangguang Zang, and Oleg \\ Knaub (cHa0s)\end{tabular} & 1.232 & \begin{tabular}[c]{@{}l@{}}Visual analytics and \\ manual inspection\end{tabular} & \begin{tabular}[c]{@{}l@{}}Raw energy meter data, \\ temporal features, building \\ metadata, simple statistical \\ features of   weather data.\end{tabular} & \begin{tabular}[c]{@{}l@{}}XGBoost, LightGBM,   \\ Catboost, and Feed-forward \\ Neural Network models \\ trained on different subset \\ of the training set\end{tabular} & \begin{tabular}[c]{@{}l@{}}Weighted mean.  \\ (different weights were used \\ for different meter types)\end{tabular} \\ \hline
3 & Xavier Capdepon & 1.234 & \begin{tabular}[c]{@{}l@{}}Eliminated 0s in the \\ same period in the \\ same site\end{tabular} & \begin{tabular}[c]{@{}l@{}}21 features   including raw data, \\ weather, and various meta data\end{tabular} & \begin{tabular}[c]{@{}l@{}}Keras CNN, LightGBM \\ and Catboost\end{tabular} & Weighted average \\ \hline
4 & Jun Yang & 1.235 & \begin{tabular}[c]{@{}l@{}}Deleted outliers during \\ the training phase\end{tabular} & \begin{tabular}[c]{@{}l@{}}23 features including raw data, \\ aggregate, weather lag \\ features, and target encoding.   \\ Features are selected using \\ sub-training sets.\end{tabular} & \begin{tabular}[c]{@{}l@{}}XGBoost (2-fold, 5-fold) \\ and Light GBM (3-fold)\end{tabular} & \begin{tabular}[c]{@{}l@{}}Ensembled three   models. \\ Weights were determined \\ using the leaked data.\end{tabular} \\ \hline
5 & \begin{tabular}[c]{@{}l@{}}Tatsuya Sano,   Minoru \\ Tomioka, and Yuta Kobayashi \\ (mma)\end{tabular} & 1.237 & \begin{tabular}[c]{@{}l@{}}Dropped long streaks \\ of constant values and \\ zero target values.\end{tabular} & \begin{tabular}[c]{@{}l@{}}Target encoding using \\ percentile and proportion and \\ used the weather data temporal \\ features\end{tabular} & \begin{tabular}[c]{@{}l@{}}LightGBM in two steps -- \\ identify model parameters on \\ a subset and then train on the \\ whole set for each building.\end{tabular} & Weighted average. \\ \hline
9 & MPWARE & 1.241 & \begin{tabular}[c]{@{}l@{}}Remove outlier,  \\ Imputation\end{tabular} & \begin{tabular}[c]{@{}l@{}}Timestamp feature, Holiday \\ feature, Categorical statistic \\ feature\end{tabular} & \begin{tabular}[c]{@{}l@{}}LightGBM, CatBoost, \\ LiteMORT, Neural Network\end{tabular} & Ensemble Model \\ \hline
13 & Tim Yee & 1.243 & \begin{tabular}[c]{@{}l@{}}Remove outlier,  \\ Imputation\end{tabular} & \begin{tabular}[c]{@{}l@{}}Timestamp feature, \\ Categorical statistic feature\end{tabular} & LightGBM & Ensemble Model \\ \hline
20 & [ods.ai]PowerRangers & 1.244 & \begin{tabular}[c]{@{}l@{}}Remove outlier,  \\ Imputation\end{tabular} & Timestamp feature & LightGBM, Neural Network & Ensemble Model \\ \hline
25 & Georgi Pamukov & 1.245 & \begin{tabular}[c]{@{}l@{}}Remove outlier,  \\ Imputation\end{tabular} & \begin{tabular}[c]{@{}l@{}}Timestamp feature, \\ Holiday feature\end{tabular} & \begin{tabular}[c]{@{}l@{}}LightGBM, Neural  Network, \\ L1/L2 regression models\end{tabular} & Ensemble Model \\ \hline
46 & Fernando Wittmann & 1.248 & - & - & LightGBM & Ensemble Model \\ \hline
\end{tabular}}
\caption{Overview of the publicly available solutions among the top 50 submissions. Only the top five winning teams were required to provide details of their solutions, but additional teams also publicly shared their solutions as Kaggle notebooks.}
\label{tab:overview_top50}
\end{table*}

The submissions are identified by a unique ID and consist of an hourly prediction per meter (2,380 meters in total). Each submission was evaluated during the competition using the \textit{Root Mean Square Log Error} (RMSLE) due to its widespread use and ability to account for differences in the magnitude of meter values: 

\begin{equation}
RMSLE = \sqrt{\frac{1}{n} \sum_{i=1}^n (\log(p_i + 1) - \log(a_i+1))^2 }
\label{eq:rmsle}
\end{equation}

Where $n$ is the total number of observations in the data set, $p_i$  is the meter reading prediction submitted by the contestant, and $a_i$ is the actual meter reading.

To summarise the top fifty submissions, the RMSLE was calculated grouping by meter, building ID, and date: one data set per kind of meter (chilled water, electricity, hot water, and steam) was obtained, containing the RMSLE value calculated from the predictions of the top 50 submissions by building and date. All negative values were removed to calculate this metric.  The RMSLE metric was scaled between 0 and 1 using the Min-Max scaler technique for a more appropriate comparison. Thus, RMSLE$_{scaled}$ is used for all the analyses presented here.

\subsection{Model error categorization}

Two types of error categories were created to characterize the reasons for the model error. These categories were created through exploratory analysis of error data as well as logical thresholds based on the usability of the model prediction outputs.

\subsubsection{Magnitude and Reach Category}

The first category is called the \emph{Magnitude and Reach (MR)} type, and it relates to the magnitude of the error and its reach amongst the buildings within a site. The concept of \emph{reach} describes whether the error impacts a single or small set of buildings or a more extensive or majority portion of the buildings from a common site. The MR types are: 

\begin{itemize}
    \item \textit{Type A - Single Building In-Range Error} - Error that occurs on a single or small set of buildings on a site with an RMSLE$_{scaled}$ threshold between 0.1 and 0.3.
    \item \textit{Type B - Multiple Building In-Range Error} - Error \textit{Type A} that occurs on at least 33\% of buildings on a site with an RMSLE$_{scaled}$ threshold between 0.1 and 0.3. This threshold was selected as a reasonable line between common behavior happening coincidentally or not.
    \item \textit{Type C - Single Building Out-of-Range Error} - Error that occurs on a single or small set of buildings on a site with an RMSLE$_{scaled}$ threshold greater than 0.3.
    \item \textit{Type D - Multiple Building Out-of-Range Error} - Error \textit{Type C} that occurs on at least 33\% of buildings on a site with an RMSLE$_{scaled}$ threshold greater than 0.3.
\end{itemize}

The RMSLE$_{scaled}$ threshold for the categories were selected to be consistent across all meter types and were based on the quartile and intraquartile ranges for the error distributions. These metrics for each meter type are found in Table \ref{tab:iqr}. Predictions from the competition that falls below the thresholds for these types (RMSLE$_{scaled}$ $<$ 0.1) is considered a \emph{good fit} as it falls within a range of relative accuracy. 

\begin{table}[]
\centering
\resizebox{0.45\textwidth}{!}{%
\begin{tabular}{|l|l|l|l|l|l|}
\hline
\textbf{Meter} & \textbf{Q1} & \textbf{Q2} & \textbf{Q3} & \textbf{IQR} & \textbf{Q3 + 1.5 x IQR} \\ \hline
Chilled Water & 0.03 & 0.06 & 0.13 & 0.10 & 0.30 \\\hline\
Electricity  & 0.01 & 0.02 & 0.04 & 0.03 & 0.21 \\\hline
Hot Water     & 0.06 & 0.14 & 0.27 & 0.20 & 0.43 \\\hline
Steam        & 0.03 & 0.06 & 0.14 & 0.11 & 0.31\\\hline
\end{tabular}%
}
\caption{Quartiles and interquartile range for the scaled RMSLE in each kind of meter.}
\label{tab:iqr}
\end{table}

\begin{figure}[t]
\begin{center}
\includegraphics[width=0.45\textwidth, trim= 0cm 0cm 0cm 0cm,clip]{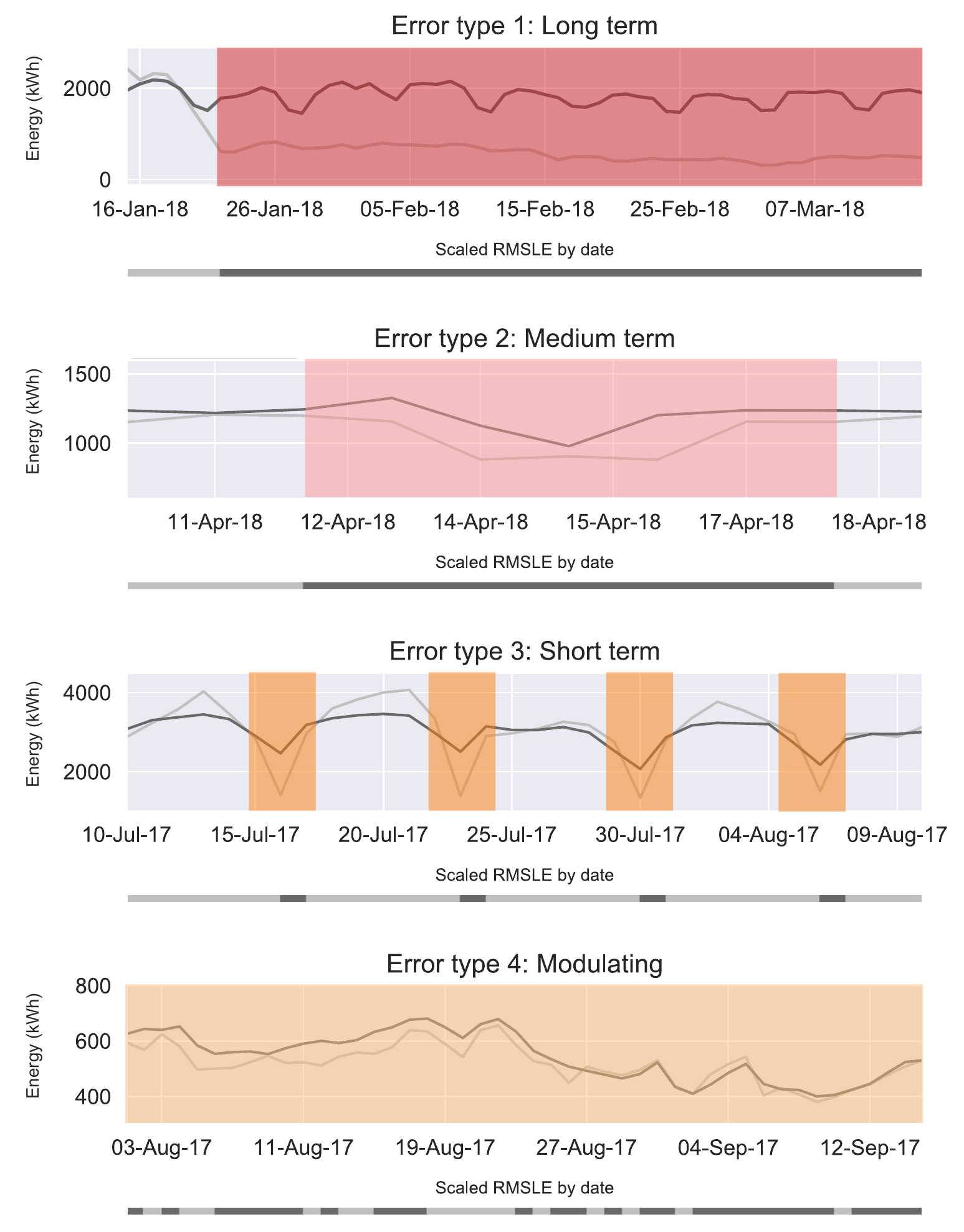}
\caption{Example of the four error types created in this analysis to characterize the temporal behavior of the error: Type 1 is a longer-term error that abruptly starts and has an extended amount of time, Type 2 and 3 are similar, yet for shorter amounts of time, and Type 4 is a fluctuating error that alternates quickly in its goodness-of-fit.}
\label{fig:error_examples}
\end{center}
\end{figure}

\subsubsection{Temporal Behavior Category}
The second category is called the \emph{Temporal Behavior (TB)} type, and it focuses on the temporal nature of the error, mostly based on the length of time the MR error persisted. The TB types are:

\begin{itemize}
    \item \textit{Type 1 - Long term error} - Long-term error likely caused by some major consistent operational change or long-term meter failure. This error is detected when the RMSLE$_{scaled}$ is over a threshold (between 0.1 and 0.3 for \textit{in-range errors} and over 0.3 for \textit{out-of-range errors}) for a time period longer than three consecutive days.
    \item \textit{Type 2 - Medium-term error} - Medium-term error with a shorter span than long-term generally from shorter seasonal effects not characterized in the training data. This error is detected when the RMSLE$_{scaled}$ is over a threshold (between 0.1 and 0.3 for \textit{in-range errors} and over 0.3 for \textit{out-of-range errors}) for a time period longer than one day but shorter than three consecutive days.
    \item \textit{Type 3 - Short-term error} - Short-term error that spans during only one day and is usually related to non-routine events and momentary meter or system failures. This error is detected when the RMSLE$_{scaled}$ is over a threshold (between 0.1 and 0.3 for \textit{in-range errors} and over 0.3 for \textit{out-of-range errors}) for one day.
    \item \textit{Type 4 - Modulating error} - Error that modulates relatively rapidly to signify that there are dynamic causes of the error. Defined as error that occurs during a \textit{w} (30-day window), and at least a proportion of them (0.1) is a short-term error (Type 3).
\end{itemize}

Figure \ref{fig:error_examples} illustrates simple examples of these four TB categories from the data set. Not all prediction time stamps will have an MR or TB category assigned to them as much of the test data is considered a \emph{good fit} if the RMSLE$_{scaled}$ is lower than 0.1. For the rest of this paper, the errors will be described with a combination of these two categories in the figures, i.e., \emph{A1} would indicate a prediction that has been assigned to MR Type A and TB Type 1.


\section{Results}
\label{sec:results}

The implementation of the screening analysis resulted in the characterization of error types across the GEPIII competition sites from the test data set. This section outlines those results and creates heat maps that illustrate each error type category's diversity and behavior patterns. Next, there is an aggregation of those errors at the site, building primary use type, and total error levels.

\subsection{Temporal error analysis}
The initial implementation of these error type filters enables the characterization of the nature of the residual errors throughout the prediction time frame. The following subsection outlines the implementation of error type filtering from both categories. Each meter type (electricity, chilled water, hot water, and steam) is visualized using a heat map with a color palette of the error type categories for both Magnitude and Reach and Temporal Behavior. Detailed information about the sites and buildings can be found in the BDG2 project and its associated publication~\citep{Miller2020-hc}.

\subsubsection{Electrical meters}
\begin{figure*}[h!]
\begin{center}
\includegraphics[width=0.8\textwidth, trim= 0cm 0cm 0cm 0cm,clip]{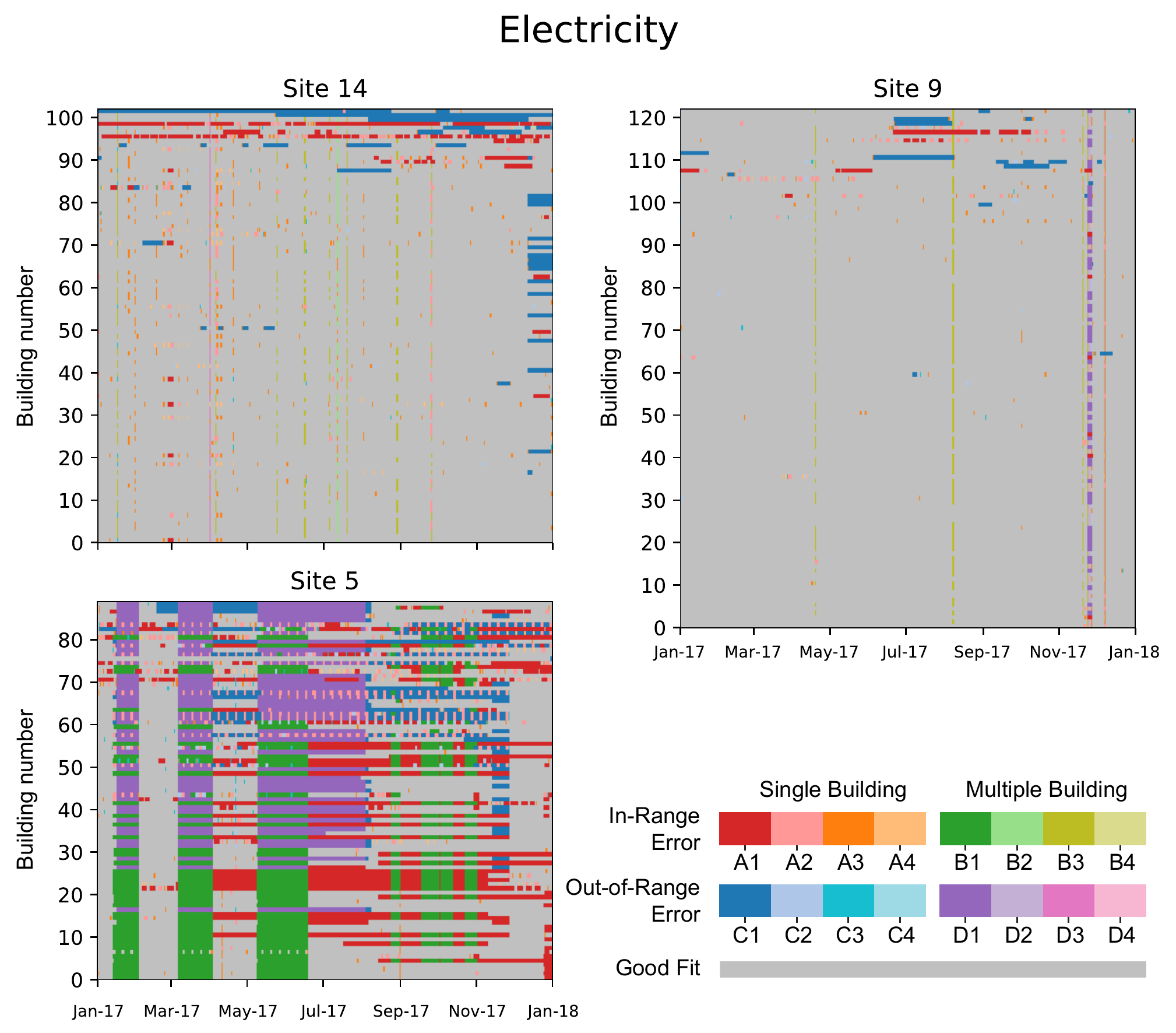}
\caption{Heat map representation of the sixteen categories of error for electricity prediction across the error type categories for Sites 5, 9, and 14. The x-axis is the time range of the public test/validation (2017) and the y-axis are the meters from that site sorted from top to bottom according to highest to lowest error prevalence.}
\label{fig:heatmap_elec}
\end{center}
\end{figure*}

The first meter type is electricity, with Sites 5, 9, and 14 analyzed. Figure \ref{fig:heatmap_elec} shows heat maps that illustrate the error type screening for three sites from the competition that are most representative of this meter. The x-axis of the heat map is the time frame of daily data for the validation/public leaderboard test data set (2017). The y-axis is the electricity meters from each site sorted from top to bottom from highest to lowest average error. This visualization technique is repeated in the upcoming figures for the other meter types.

Site 5 shows a significant amount of error that seems to be related to seasonal scheduling across the entire site. There are large clusters of consistent single and multiple building errors (Type A/B) with consistent cut-offs and seem to follow a particular phase-based schedule. Most of these errors are also perceived to be long-term (Type 1). In addition, there are several clusters of out-of-range (Type C/D) errors that overlap with the periods on several buildings. Site 5 is made up mostly of municipal buildings, and many of the buildings are primary and secondary schools that have set schedules where classes are likely in or out-of-session~\citep{Miller2020-hc}.

For sites 9 and 14, there are much fewer errors types found for most of the meters, with the majority being either individual buildings (Type A/C) that have frequent bursts of error (towards the top of the heat maps) or short time frames of errors occurring across most of the buildings on the site at the same time. These sites are university campuses, and their errors seem to show that most electrical consumption behavior is relatively easy to predict except for short and sometimes systematic exceptions~\citep{Miller2020-hc}. A contribution to this ability to more accurately predict consumption might also be the higher quality of data produced by potentially revenue-generating electricity meters.

\begin{figure*}[h!]
\begin{center}
\includegraphics[width=0.85\textwidth, trim= 0cm 0cm 0cm 0cm,clip]{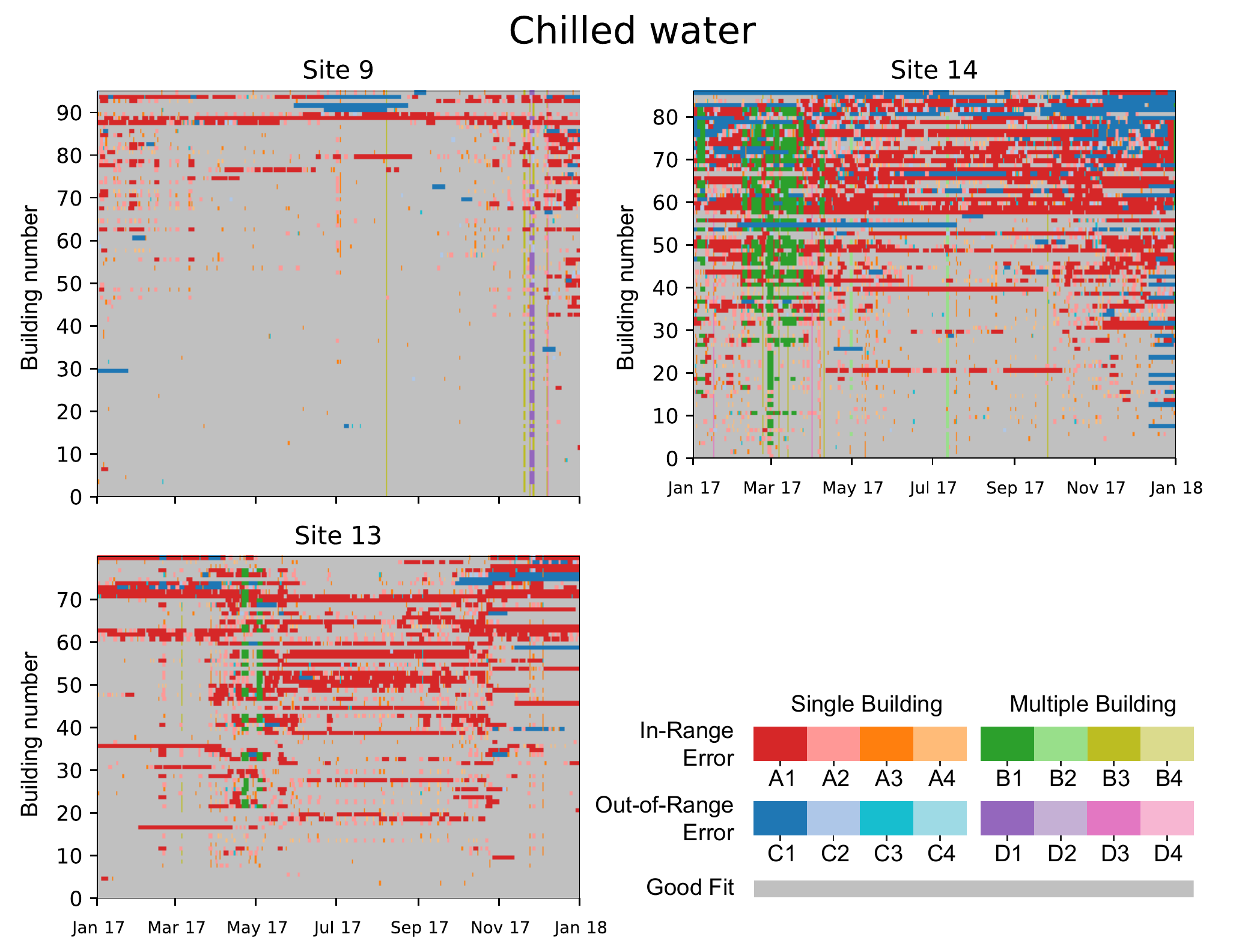}
\caption{Heat map representation of the sixteen categories of error for chilled water prediction across the error type categories for Sites 9, 13, and 14. The x-axis is the time range of the public test/validation (2017) and the y-axis are the meters from that site sorted from top to bottom according to highest to lowest error prevalence.}
\label{fig:heatmap_chw}
\end{center}
\end{figure*}

\begin{figure}[t!]
\begin{center}
\includegraphics[width=0.45\textwidth, trim= 0cm 0cm 0cm 0cm,clip]{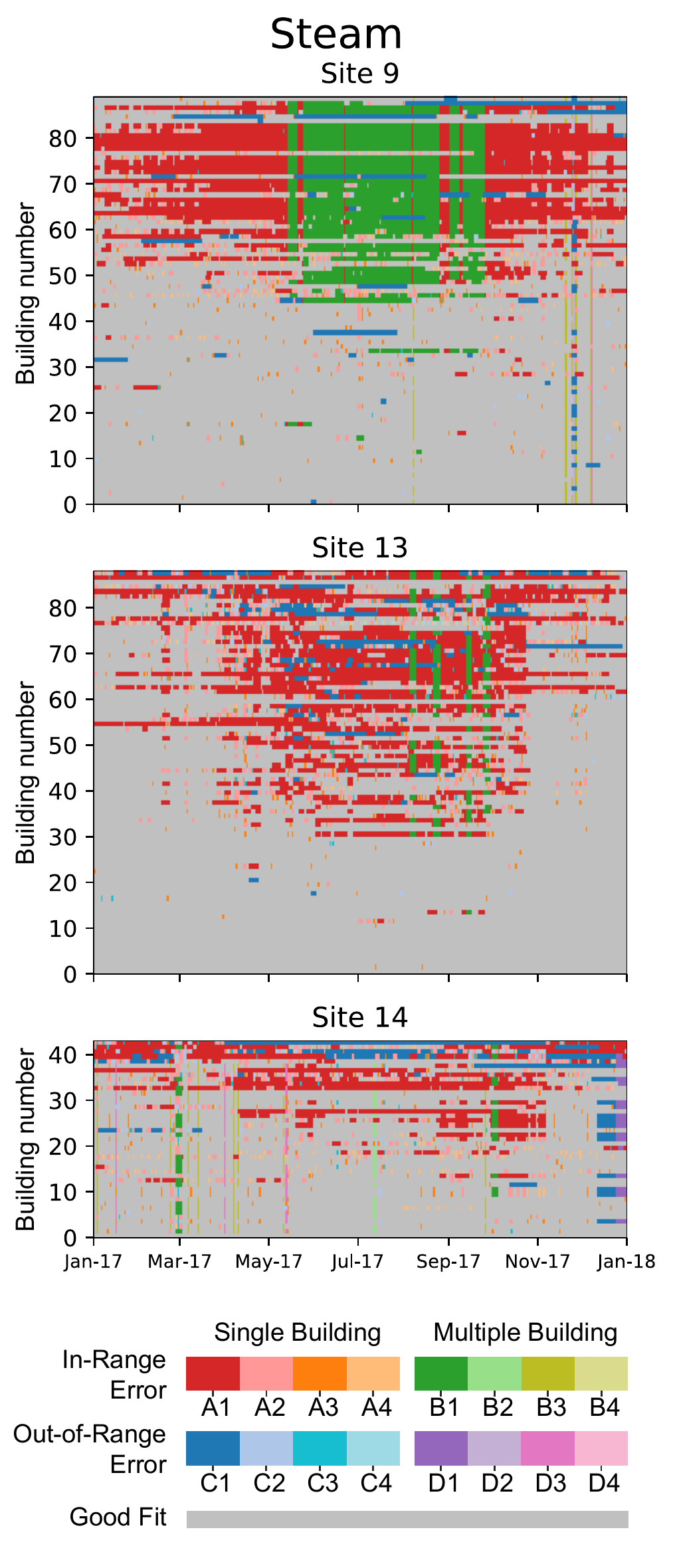}
\caption{Heat map representation of the sixteen categories of error for steam prediction across the error type categories for Sites 9, 13, and 14. The x-axis is the time range of the public test/validation (2017) and the y-axis are the meters from that site sorted from top to bottom according to highest to lowest error prevalence.}
\label{fig:heatmap_steam}
\end{center}
\end{figure}

\subsubsection{Chilled water meters}
Figure \ref{fig:heatmap_chw} illustrates the screening heat maps for Sites 9, 13, and 14 for the public test/validation data set. Once again, these three sites were chosen based on their representation of this meter type. The visualization from these three sites quickly shows much more error occurring than with electricity. All three of these sites are university campuses with chilled water plants serving various buildings. Most errors are long-term (Type 1) and come from single buildings (Type A/C). There seems to be some heating and cooling seasonality in the errors, with some of the meters distinctly having more error in the winter and spring months, while other buildings have more error in the summer and fall months. There is less uniformity in the starting and stopping errors across buildings despite large groups of buildings seeming to experience similar errors simultaneously. There are variations in error magnitude, with most errors being in-range (Type A/B) with pockets of out-of-range (Type C/D) sometimes occurring in clusters.

\subsubsection{Steam meters}
Figure \ref{fig:heatmap_steam} shows the error type behavior for the same three sites (9, 13, and 14) as the chilled water. All three sites show approximately half of the meters have good predictions with few errors. The other half has a significant amount of long-term (Type 1) error, especially during the cooling season (for Sites 9 and 13). The steam system energy consumption is driven by behavior different from the previous year and not by weather differences. These sites are all university campuses, and the control of these systems seems to be more sporadic than electricity and chilled water meters. 

\subsubsection{Hot water meters}
\begin{figure}[h!]
\begin{center}
\includegraphics[width=0.45\textwidth, trim= 0cm 0cm 0cm 0cm,clip]{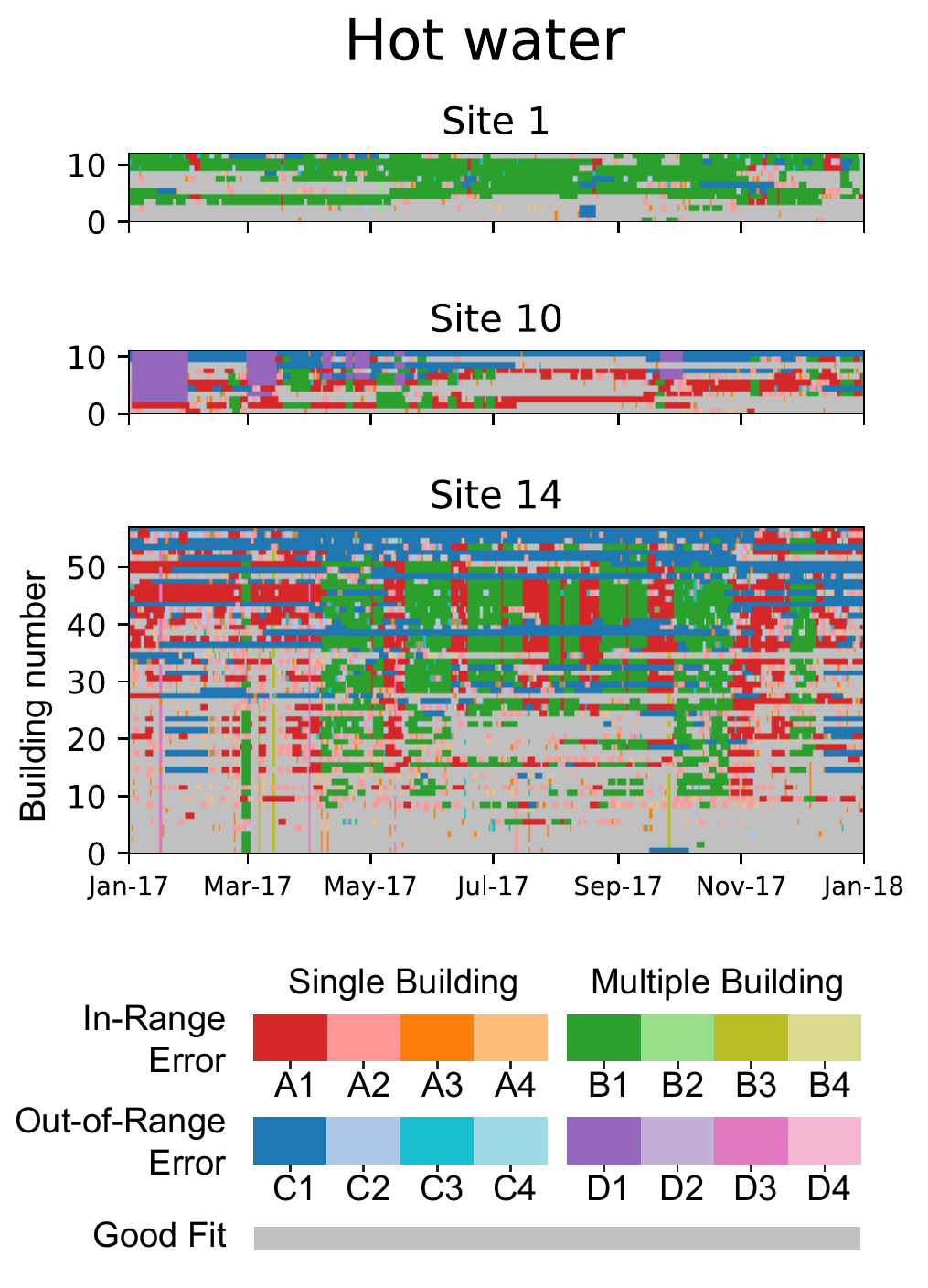}
\caption{Heat map representation of the sixteen categories of error for hot water prediction across the error type categories for Sites 1, 10, and 14. The x-axis is the time range of the public test/validation (2017) and the y-axis are the meters from that site sorted from top to bottom according to highest to lowest error prevalence.}
\label{fig:heatmap_hw}
\end{center}
\end{figure}

Finally, Figure \ref{fig:heatmap_hw} gives an overview of three representative sites of hot water meters. Sites 1, 10, and 14 are covered in this visualization and are all university campuses with some buildings on a centralized hot water supply network. Initially, it is evident that the number of hot water meters is lower than other types, and they seem to have much more error across a broader range than electricity or chilled water. An interesting situation in Site 10 occurs when the entire first month of the data is entirely out-of-range across all buildings. This situation could be an example of total system failure of either the data collection system or the meters themselves. Site 14 shows consistent error across most buildings, with bands of out-of-range errors across many buildings during similar periods.

\subsection{Breakdown of error types across sites and meter types}
\begin{figure*}[]
\begin{center}
\includegraphics[width=0.88\textwidth, trim= 0cm 0cm 0cm 0cm,clip]{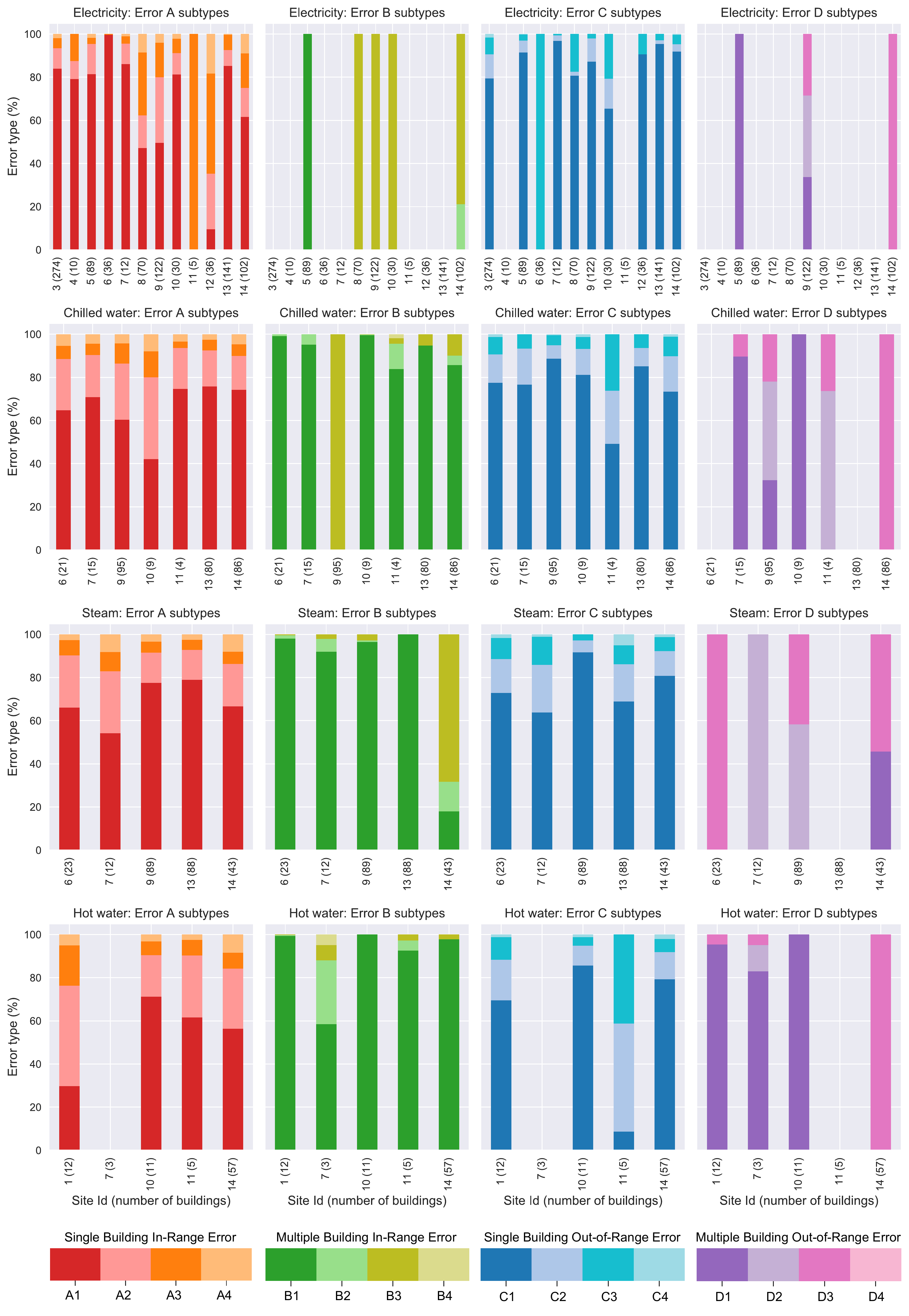}
\caption{Aggregations of the Temporal Behaviour (Types 1-4) across the Magnitude and Reach errors (Types A-D), sites, and meter categories. The number of buildings in each meter type category is in parentheses next to the x-axis label. If the bar is missing for a particular error, then that category wasn't detected for that site.}
\label{fig:subtypes_breakdown}
\end{center}
\end{figure*}

Figure \ref{fig:subtypes_breakdown} illustrates the breakdown of Types 1-4 across the Magnitude and Reach categories (Types A-D) for all sites and meter types. The only error present in all sites for electricity meters is the single building in-range error (Type A). Multiple building errors (Types B/D) are only present in Sites 5 and 3, respectively. Out-of-range single building error (Type C) is present in almost all sites. For single building errors (Types A/C), the long-term category (Type 1) seems to be the majority in contrast to multiple building errors (Types B/D), where there is more diversity. All error types except a few multiple building out-of-range errors are found in all chilled water sites. Chilled water meters also have a significant long-term error (Type 1), similar to electricity. Similar behavior is observed for the steam meter. In the case of hot water meters, only multiple building in-range error (Type B) is present in all sites. In all cases, multiple buildings out-of-range error (Type D) seems less frequent but the most diverse in Types 1-3. It can be noticed that multiple building out-of-range error (Type D4) was not detected in the experimental results.

\subsection{Aggregated error analysis across sites}
\begin{figure*}[]
\begin{center}
\includegraphics[width=0.88\textwidth, trim= 0cm 0cm 0cm 0cm,clip]{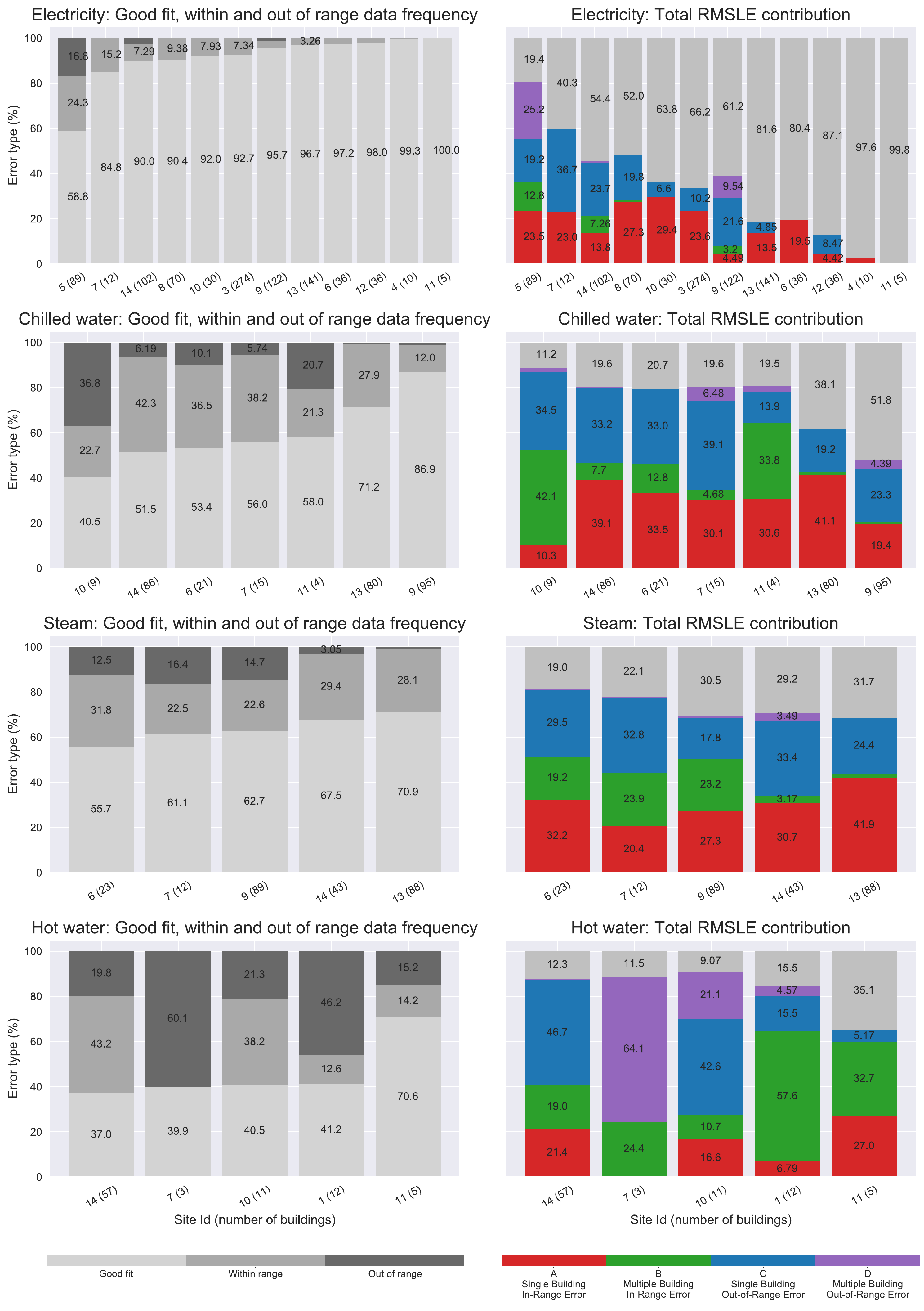}
\caption{Aggregations of error types across the meter types (from top to bottom: Electricity, Chilled Water, Hot Water, and Steam) and sites with percentages of in-range (Types A/B), out-of-range (Types C,D) and Good Fit (left) and across all four Magnitude and Reach categories (Types A-D) for each site (right). The number of buildings in each meter type category is in parentheses next to the x-axis label.}
\label{fig:aggregated_sites}
\end{center}
\end{figure*}

Figure \ref{fig:aggregated_sites} shows the breakdown across all sites and meter types. This chart shows the diversity of error types across different locations. These aggregations are where it can first be seen that electricity is better predicted in general than the other meter types. The \emph{good fit} range for electricity across all sites ranges from 60-99\%. This chart also illustrates the RMSLE$_{scaled}$ contribution of each error type. Steam meters have the next best level of performance, with the \emph{good fit} category occurring 55-70\% of the time out of the five sites compared in the graphic. Most of the errors were in the in-range categories (Types A/B), with several of the sites (Sites 6, 7, and 9) also having significant single building out-of-range error (Type C). Chilled water models perform worse on average than electricity and steam, with around 40-85\% of the test data for each of the sites falling into the \emph{good fit} range. The sites with higher error, like Site 10, tend to have much more multi-building and out-of-range errors. Site 9 has the lowest error due to its relatively low amount of out-of-range error. This insight is reinforced from the heat map in Figure \ref{fig:heatmap_chw}. Hot water meters performed the worst out of the meter types. For all sites, most errors are in-range, except Site 7, which has out-of-range errors exclusively.

\subsection{Aggregated error across meter types}
\begin{figure*}
\begin{center}
\includegraphics[width=0.9\textwidth, trim= 0cm 0cm 0cm 0cm,clip]{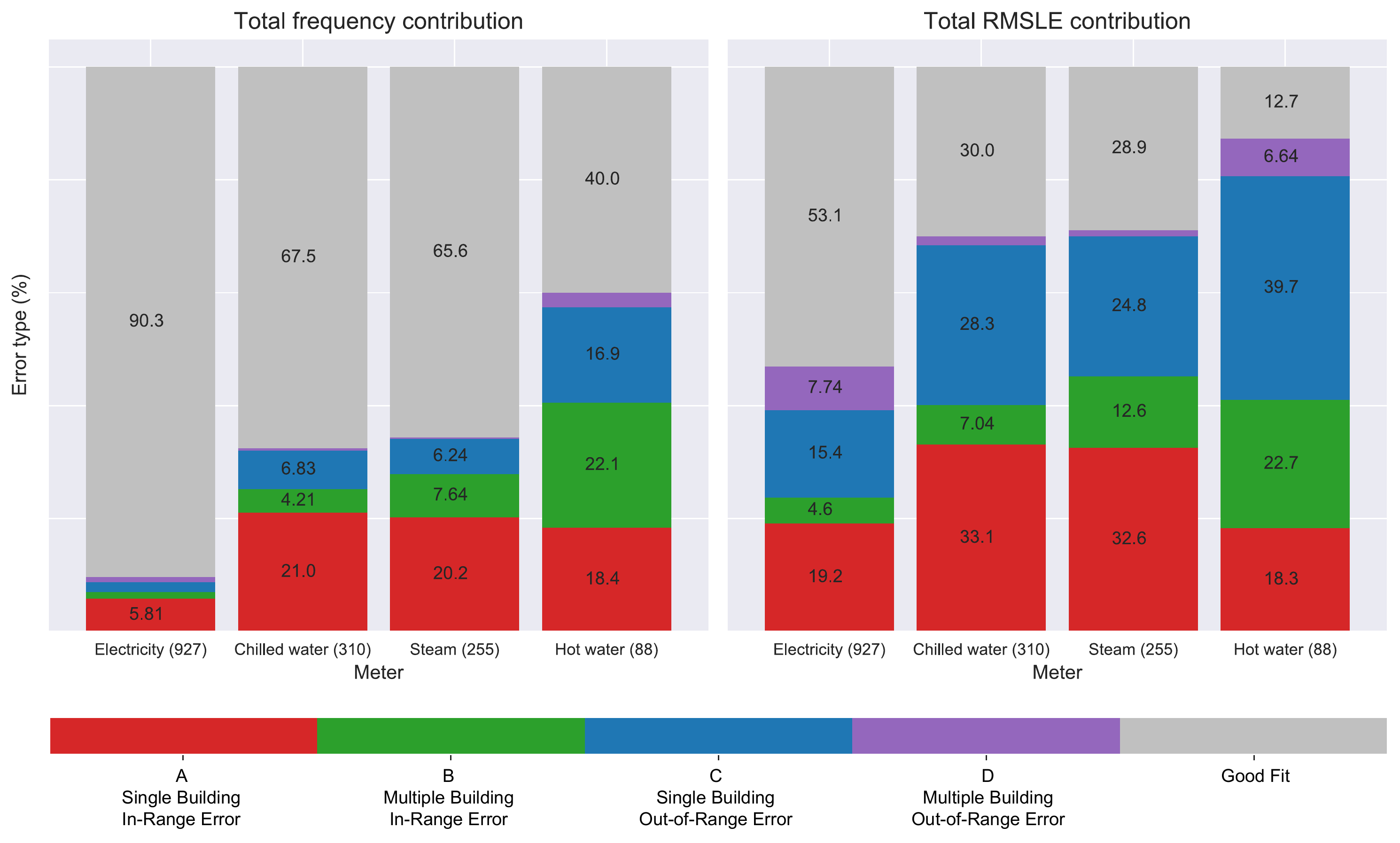}
\caption{Error breakdown (Types A-D) across the meter types including the aggregation of frequency of error types as a percentage of all errors (left) and percentage of contribution to the scaled RMSLE error for the meter across the entire test data set. The number of buildings in each meter type category is in parentheses next to the x-axis label.}
\label{fig:error_summary}
\end{center}
\end{figure*}

The error analysis can be further aggregated according to the magnitude and site-related error type (Types A-D). Figure \ref{fig:error_summary} shows this breakdown according to its impact on both the frequency of error occurrence and the RMSLE$_{scaled}$ contribution for the whole test data set. Among all meter types, electricity meters have the best prediction accuracy (90.3\% good fit), chilled water and steam meters have the next best accuracy (65.6 and 67.5\% good fit), and hot water meters models the least accurate with only 40.0\% of the predictions fall into the \emph{good fit} category. Regarding the distribution of RMSLE$_{scaled}$ contribution in each meter type, most of the error contributions come from a single building (Type A/C), significantly more than those of multiple buildings (Type B/D). Only the electricity meters have more than half of the errors contributed by the good-fit category.

The electricity meter has the best prediction accuracy, with more than half of the frequency and RMSLE$_{scaled}$ contributions in the good-fit category. The high contribution of single building error (Types A/C) indicates that the errors are mainly from individual buildings rather than cross-building events. As for chilled water and steam meters, which have very similar distributions, their prediction errors are significantly higher than that of the electricity meter, with more than 50\% of the errors coming from single buildings (Types A/C). This situation shows the greater difficulty in predicting these two meter types and points out that the source of error mainly comes from individual buildings. 

The hot water meters with poor prediction performance also have more errors from single rather than from multiple buildings. However, the in-range error of numerous buildings (Type B) is higher than the in-range error of a single building (Type A). This situation indicates that the hot water meters have more errors due to systematic events across a campus. For example, for Site 14 of Figure \ref{fig:heatmap_steam}, consistent in-range errors across buildings co-occur during specific time periods.

\begin{figure*}[]
\begin{center}
\includegraphics[width=0.9\textwidth, trim= 0cm 0cm 0cm 0cm,clip]{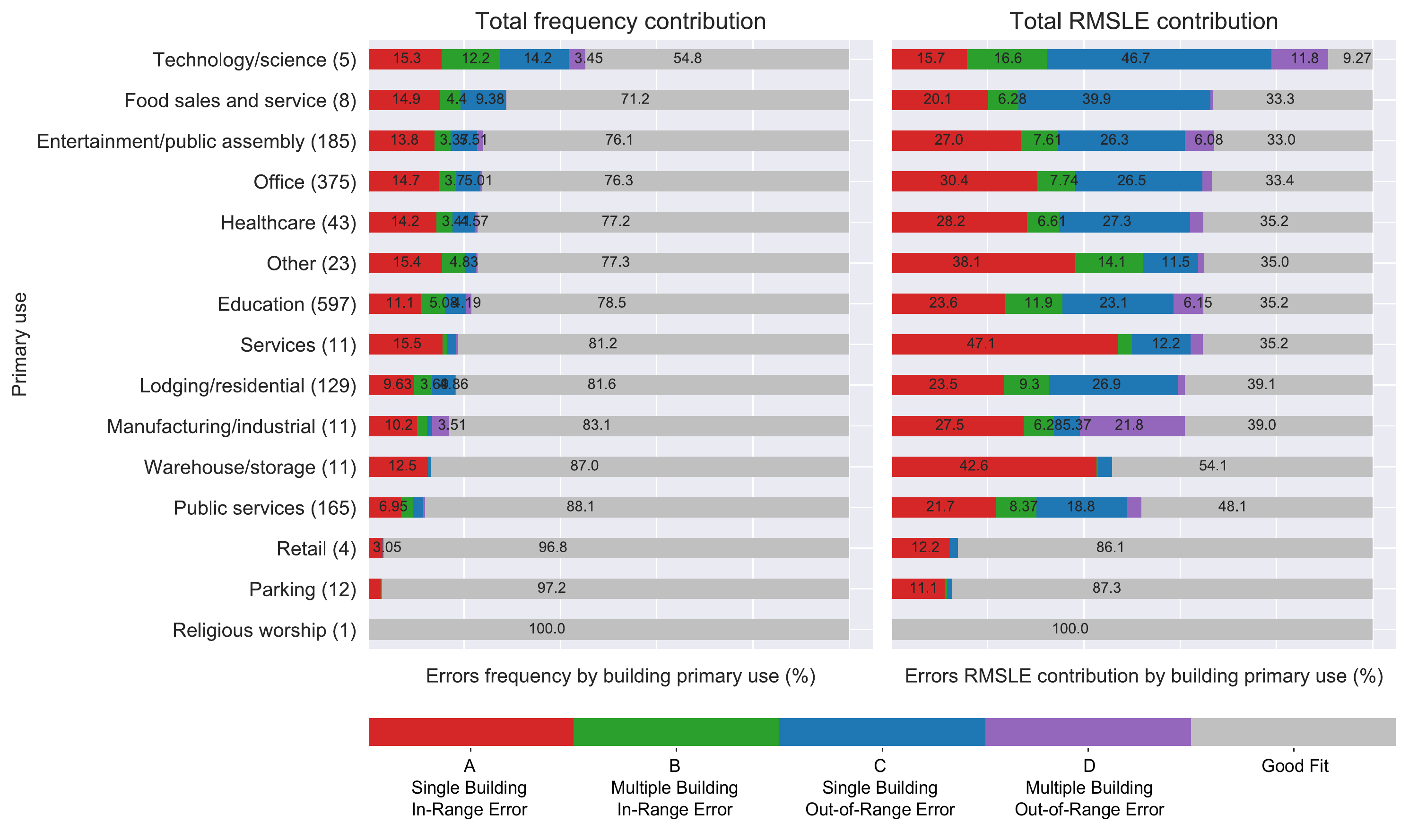}
\caption{Error breakdown (Types A-D) from all meter types across the various primary use types of the buildings which the meters served. The number of buildings in each category is in parentheses next to the y-axis label}
\label{fig:primary_use_summary}
\end{center}
\end{figure*}

\subsection{Aggregated error across building primary use types}
Figure \ref{fig:primary_use_summary} illustrates the Magnitude and Reach errors (Types A-D) as broken down by building primary use type. Most errors come from single buildings (Types A/C), and fewer errors come from cross-building events (Types B/D). In addition, except for some building types with only a few samples (e.g., Technology/Science, Food sales, and Service), almost all building types have good-fit categories covering 76.1 - 100.0\% of error frequency. 

Regarding the error contribution in terms of RMSLE$_{scaled}$ for the major building types, Entertainment/public Assembly, Office, and Lodging/residential all have a reasonably similar contribution structure, with the majority of the error coming from single compared to multiple buildings. Interestingly, Education and Technology/science buildings have a higher proportion of multi-building within-range errors (Type B), accounting for 11.9\% and 16.6\% of the RMSLE$_{scaled}$. The possible reason for this situation could be that these buildings are primarily from university campuses and have different energy use behavior during semesters and vacations. In this competition, because the prediction model lacks site-specific schedule features, there are more simultaneous prediction errors for multiple buildings during holidays or break periods.

\subsection{Overall error breakdown analysis}
\begin{figure*}[]
\begin{center}
\includegraphics[width=0.85\textwidth, trim= 0cm 0cm 0cm 0cm,clip]{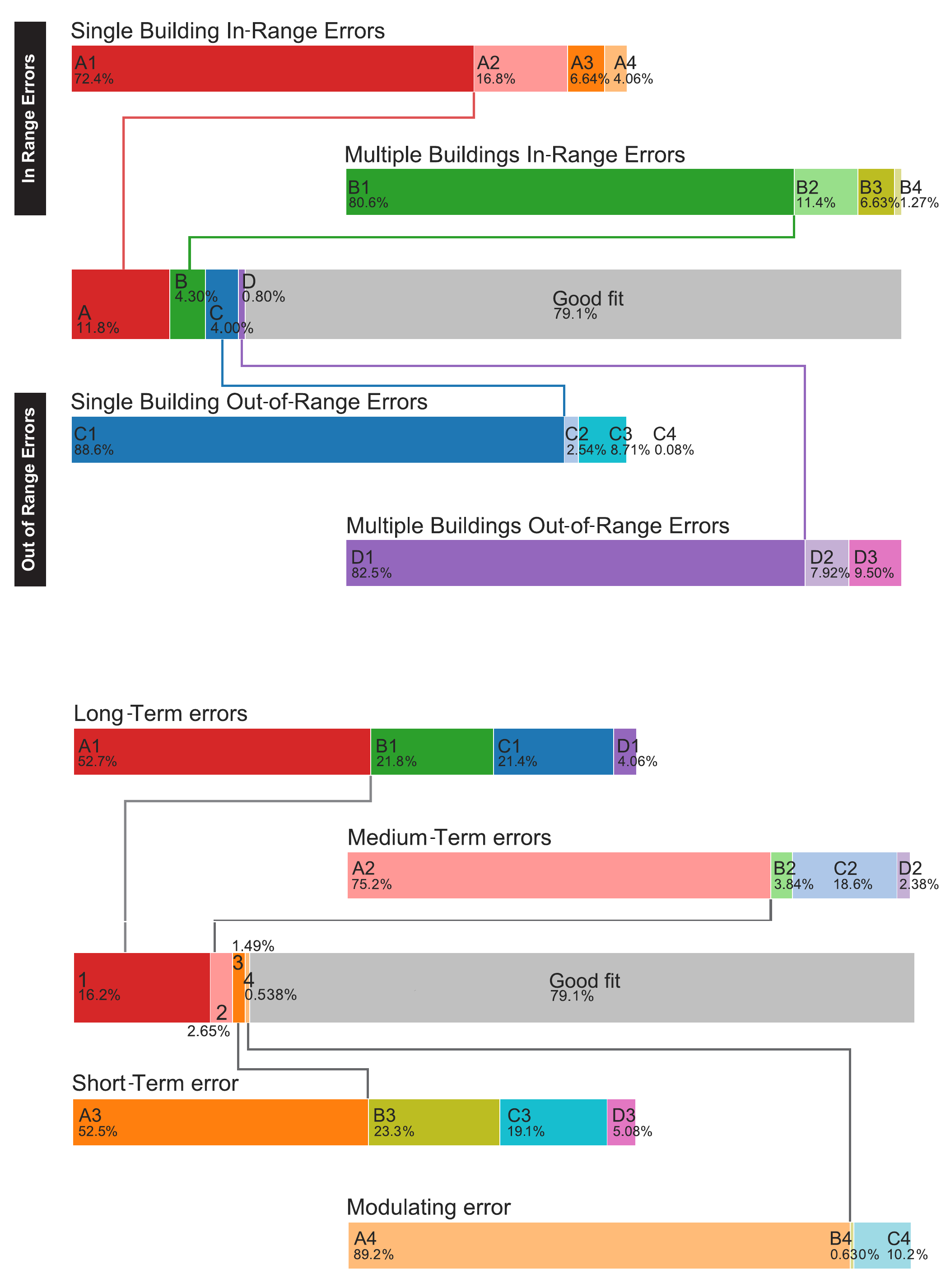}
\caption{Highest level overview of errors with the Magnitude and Reach categories (Types A-D) breakdown shown in the middle and the temporal behaviour type breakdowns (Types 1-4) according to in-range (top) and out-of-range (bottom) errors.}
\label{fig:total_error_breakdown}
\end{center}
\end{figure*}

The highest level aggregation for this paper is shown in Figure \ref{fig:total_error_breakdown}, which illustrates the breakdown of Magnitude and Reach (MR) categories and a breakdown of each according to the Temporal Behavior categories (Types 1-4).  This aggregation gives a high-level understanding of the proportion of different errors from the GEPIII competition. The in-range, single building error (Type A) accumulates to 11.8\% of the time frame from the test data set while the multi-building version (Type B) is at 4.3\%. This total of 16.1\% is the errors from the meters that fall within a range that has the potential to be \emph{fixed} through alternative data sources or other innovations. Out-of-range error (Types C/D) makes up 4.8\% of the total time frame from the test data set. These errors can be considered more extreme in their magnitude and are likely due to unpredictable non-routine events.

\section{Discussion}
\label{sec:discussion}

The results of this study provide a foundation for suggested methods to increase the accuracy of building energy prediction. The following sub-sections outline strategies and technologies that address the error type definitions from the results. 

\subsection{Addressing in-range errors from a single building by incorporating sensors and other new data feeds}
The single building in-range error classification (Type A) is an example of when a single or small subset of buildings have long, medium, short, or modulating error behavior that falls within a specific range of magnitude. These periods of less severe error theoretically are caused by phenomena not captured by previous training data related to systematic human behavior at the building level. Addressing these errors is likely a matter of finding other temporal data sources that could be harvested and used in the training/testing data. 

For example, carbon dioxide sensors from the building management system could be used as a proxy for occupancy in a building whose energy consumption is sensitive to internal loads from people~\citep{Miller2021-bm}. This signal could reduce error, especially for A3 and A4 errors that are shorter in length or more modulating in nature. Several previous studies have incorporated such data types for the purpose of subsystem optimization for cooling, heating, and systems optimization~\citep{Platon2015-jd}. 

Long and medium-term errors (Types A1/A2) are likely more a result of a system or environmental changes that are not captured in previous training data and are not weather influence. These are situations when zone air temperature setpoints are changed or when systems are turned down or off compared to previous behavior. These errors could be mitigated by incorporating these signals from the building management system (BMS) into the training and test data sets. The characterization of these types of patterns has momentum for systems optimization~\citep{Capozzoli2017-dk} and model calibration~\citep{Chong2021-hs} and further work in this direction would benefit multiple fields.

\subsection{Addressing in-range errors for multiple buildings from the same location using site-wide data to detect systematic disruptions}
The next category of errors to discuss is those where the residual is present across numerous buildings at the same site (Type B). This scenario indicates abnormal energy-consuming behavior at a larger scale, such as district-scale occupancy, climate, or other types of events or failures. These errors are likely also due to either change in human behavior or system control but at a different scale. Similar to the single building errors (Type A), there are short-term and modulating increases in residual due to site-wide human behavior (Types B3/B4). This scaled-up impact is likely due to scheduling scenarios where there is a site-wide change in building use. An example of this situation is when a university campus has a break period or holiday that only exists at that specific site (and did not happen precisely the same way in the training data). Several recent works have utilized various sensors at the campus level for the development of occupancy patterns~\citep{Ding2021-al}, systems control~\citep{Liu2021-pw}, and indoor air quality prediction~\citep{Tagliabue2021-wr}. Mitigation of this error might be possible with additional data from online calendars regarding the seasonal changes in the use of buildings. Potential online or third-party data sources that include signals relevant to energy use might be used to mitigate this error, including from social media~\citep{Lu2021-xm, Fu2022-xy}. For long and medium-term errors (Types B1/B2), these could also result from system-based changes that occur in those time ranges, but this time also at the systems level. Mitigation of these errors could include data from centralized chilled water plants or supervisory control systems that may make changes at the whole site level.


\subsection{Non-routine event detection to find out-of-range errors that are unlikely to be fixed}
The error categories considered out-of-range (Type C/D) mirror the in-range (Type A/B) in all ways except the magnitude. The magnitude of these categories is such that the predictions diverge from the actual signal in these situations so much that it is unlikely to assume that just adding another training data stream or modifying machine learning parameters would be adequate. A prominent source of this type of error would be building or system sensor failures that cause the meter reading to go to zero or flat-line at a specific value. Other causes of these errors could be drastic changes in the way buildings operate due to renovation, retrofit, or significant system change. A good strategy for dealing with these errors might be to create a change-point detection trigger that essentially shows that the model is not helpful anymore~\citep{Touzani2019-sz}. There is little work in change-point event detection for building performance models, which could be a fruitful direction for enhancing modeling efforts.

\subsection{Using occupant-centric operations data in building energy prediction}
The subsections in this discussion have explored various sensors, but there are emerging opportunities to include human-related data streams into the energy prediction flow. In the literature, there are only a few instances of training data using occupants counts~\citep{Liu2013-ld}. Previous work has shown that the way occupants use buildings is different from expected~\citep{Park2019-rn, Quintana2021-zd} and data from Wifi signals of occupants~\citep{Nweye2020-yz, Zhan2021-jy}, Bluetooth localization~\citep{Tekler2020-it, Jayathissa2020-pv, Rahaman2019-qd}, and even text mining~\citep{Miller2019-if} can capture signals that could be used as inputs in the modeling process. This type of modeling approach may also influence occupant-centric controls~\citep{Gunay2021-ly}. There is significant potential in characterizing human activity in buildings for numerous research communities, and this direction should be further pursued.

\subsection{Limitations}
While this analysis is the largest of its kind for building energy prediction, it does not include a critical mass of all building types from all climate and geographic contexts worldwide. It can be seen from the open data set released from this competition that five primary use types are included in the data set: Education (classrooms), Offices, Entertainment/Public Assembly, Lodging/Residential, and Public Services. In addition, these buildings are all from North America, the UK, or Ireland~\citep{Miller2020-hc}. Each of these building use types has enough of a population of buildings so that the results can be considered generalizable for these use types and contexts, but further benchmarking work needs to be done to create a broader analysis from a more extensive and more comprehensive data set. The real challenge in addressing these limitations is the data availability from actual buildings, and building owners now have a low incentive to capture or share data with the research community. This work is a first step in gathering ever-larger and more diverse data sets that the community can use to apply emerging tools and techniques going forward.

Another limitation is the sole use of the RMSLE metric in characterizing the errors in the competition. This metric is used exclusively in this analysis to make a solid link to the competition and the broader machine learning community. The building performance community relies heavily on the coefficient of variation of the root mean square error (CVRMSE) and the mean bias error (MBE)~\citep{Raftery2011-cu}. These metrics will be applied in future work using these error data focused on practical applications unrelated to the competition analysis. 

\section{Conclusions}
\label{ch:conclusion}
This paper analyzes the residual errors from the largest machine learning competition ever held for the building energy prediction domain. The results show that 79.1\% of the test data set in the competition was predicted with reasonable levels of error (\emph{good fit}). Still, there was a sizeable limitation in prediction ability for steam and hot water meters and buildings primarily used for technology, science, and food sales. Most of the regions of error were classified as being \emph{in-range} (16.1\% of the test data set) and therefore potentially addressable using alternative data sources as training data beyond just weather and metadata factors. A minority of the test data (4.9\%) were found to be from the \emph{out-of-range} error category, which is considered to be errors greater than what is reasonable to assume can be fixed within the model. These error regions likely need to be addressed using unsupervised change-point models to detect which identify when the signal is so far out of range as reasonable. 

In addition to describing the limitations of models in capturing behavior, this analysis also suggested various methods to assist the machine learning community in improving errors in this domain. For example, incorporating alternative data sources from occupancy-related sensors could reduce the error caused by on-demand energy uses in building influenced by the variability of occupancy. The addition of district-scale data such as site-specific schedules could be helpful in reducing systematic errors across collections of related buildings. And the act of detecting simply when a model is not relevant any longer due is useful in detecting fundamental shifts in the way a building uses energy, such as major retrofit, disruption, or natural disaster. These insights provide a foundation for future work in building energy prediction using data-driven methods. The limitations of this analysis can help facilitate future machine learning competitions that a) include an even larger and more diverse data set, b) provide a broader range of building use types and locations and c) incentivize building owners to donate further open data sets for community use and benchmarking. 

\subsection{Reproducibility}
Segments of the raw data and analysis code used for this study are available in an open-access Github repository that includes further documentation: \url{https://github.com/buds-lab/ashrae-great-energy-predictor-3-error-analysis}. 

\section*{CRediT author statement}
\textbf{Clayton Miller}: Conceptualization, Methodology, Formal analysis, Writing - Original Draft, Supervision, Project administration, Funding acquisition; \textbf{Bianca Picchetti}: Methodology, Software, Formal analysis, Investigation, Writing - Review \& Editing, Visualization; \textbf{Chun Fu}: Formal analysis, Data Curation, Writing - Review \& Editing; \textbf{Jovan Pantelic}: Methodology, Writing - Review \& Editing.

\section*{Funding}
The Singapore Ministry of Education (MOE) provided support for the development and implementation of this research through the \emph{Temporal Mining of Energy and Indoor Environmental Quality Data from Buildings} (R296000181133) Project.

\section*{Acknowledgements}
This analysis is possible due to the GEPIII competition planning and operations committees. The technical committee to be acknowledged includes (alphabetical order) Anjukan Kathirgamanathan, June Young Park, Pandarasamy Arjunan, and Zoltan Nagy. Planning committee members to be acknowledged are Anthony Fontanini, Chris Balbach, Jeff Haberl, and Krishnan Gowri. The ASHRAE organization is recognized for providing support for the competition prize money and the Kaggle platform for hosting GEPIII as a non-profit competition. In addition, the authors would like to thank those who assisted in collecting and releasing the BDG2 data set, including Brodie Hobson, Forrest Meggers, Paul Raftery, and Zixiao Shi.


\bibliographystyle{model1-num-names}
\bibliography{references}

\end{document}